\documentclass[letterpaper, 10pt, conference]{ieeeconf}        

\IEEEoverridecommandlockouts                              

\usepackage{graphicx} 
\usepackage{mathptmx} 
\usepackage{times} 
\usepackage{amsmath} 
\usepackage{amssymb}  
\usepackage{subcaption} 
\usepackage{booktabs} 
\usepackage{multirow} 
\usepackage[bookmarks=false]{hyperref} 
\usepackage{authblk} 
\usepackage{hyperref}

\graphicspath{{./figures/}}

\title{\LARGE \bf
Exploiting Event Cameras for \protect\\ Spatio-Temporal Prediction of Fast-Changing Trajectories
}

\author[1,2]{ Marco Monforte}
\author[2]{ Ander Arriandiaga}
\author[2]{ Arren Glover}
\author[2]{ Chiara Bartolozzi}

\affil[1]{ Universit\`a degli Studi di Genoa, Italy}
\affil[2]{ Event-Driven Perception for Robotics, Istituto Italiano di Tecnologia, Italy}
\affil[ ]{Email: \{marco.monforte, ander.arriandiaga, arren.glover, chiara.bartolozzi\}@iit.it}

\begin{document}
\maketitle
\thispagestyle{empty}
\pagestyle{empty}


\begin{abstract}
This paper investigates trajectory prediction for robotics, to improve the interaction of robots with moving targets, such as catching a bouncing ball. Unexpected, highly-non-linear trajectories cannot easily be predicted with regression-based fitting procedures, therefore we apply state of the art machine learning, specifically based on Long-Short Term Memory (LSTM) architectures. In addition, fast moving targets are better sensed using \textit{event cameras}, which produce an asynchronous output triggered by spatial change, rather than at fixed temporal intervals as with traditional cameras. We investigate how LSTM models can be adapted for event camera data, and in particular look at the benefit of using asynchronously sampled data.

\end{abstract}


\section{Introduction}
\label{sec:intro}

Predictive capabilities can massively improve the safety and reliability of robots as they interact with the environment. Humans make use of predictive estimates daily, and almost in an unconscious manner, for instance when driving, playing sports or interacting with other people. A similar capability in robots could further allow them to be integrated amongst humans, and be utilised for a wider range of tasks. In addition, forecasting information about the surrounding environment can allow a robot to plan actions in advance, and in a way to compensate for insufficiency in their actuation and movement capabilities. As such, high performance robots will not be limited to robots with high-quality, highly-precise, but costly, actuators and parts.
However, the motion of many things in the environment is not easily predictable with simple extrapolation, as, for example, biological motion can be highly non-linear and sporadic. Patterns still exist in the data, but a simple model may not be sufficient, with the need to resort to data driven learning.
Training neural networks to learn to predict trajectories has been successful in gait stability for fall-risk monitoring~\cite{gait-stability-prediction}, as well as for robot manipulator control~\cite{collaborative-hr-handover-lstm} and path planning~\cite{lstm-path-planning}.

\begin{figure}[!t]
  \centering
        \begin{subfigure}[!tb]{.49\linewidth}
            \centering
            \includegraphics[width=\linewidth]{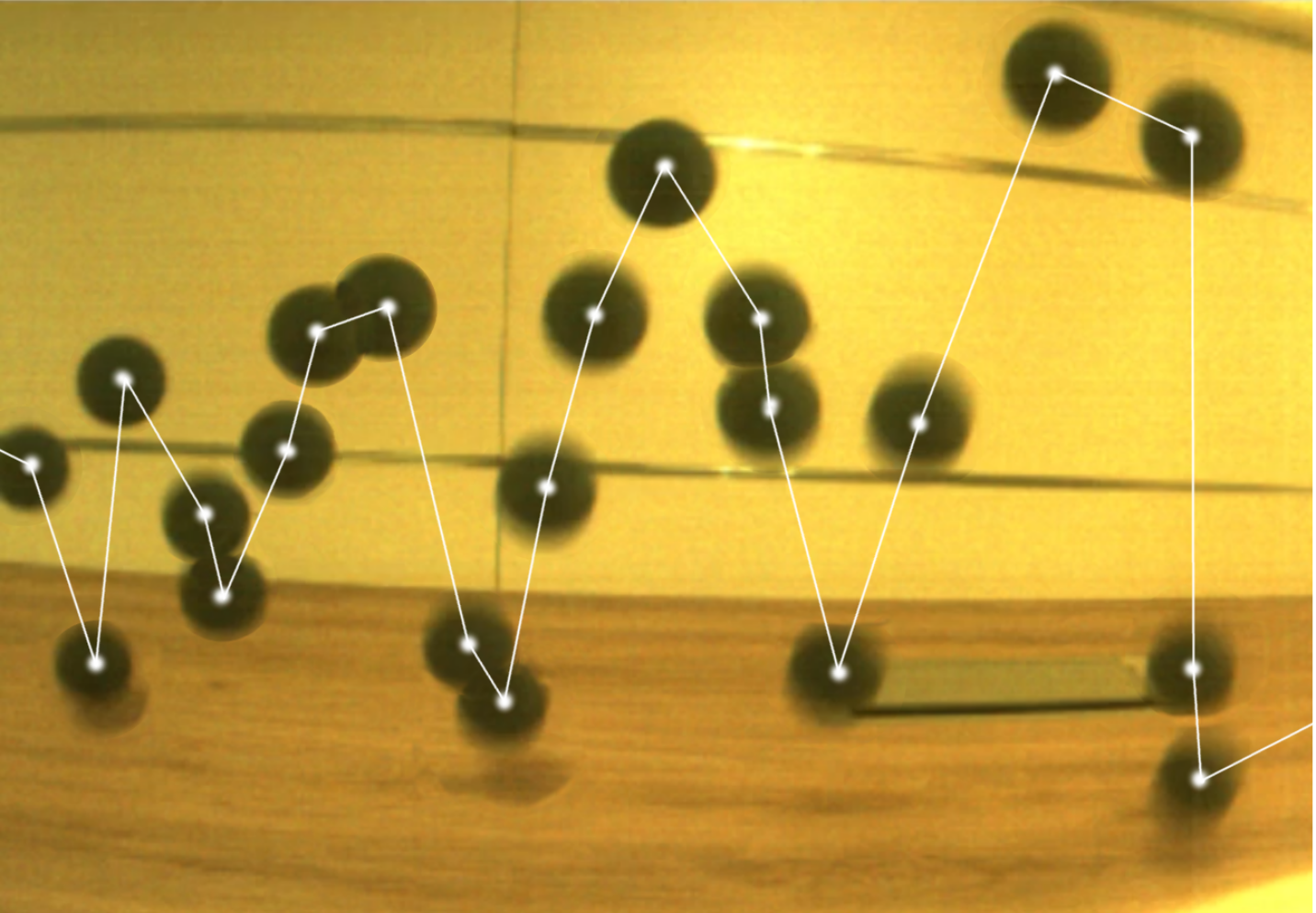}
            \caption{}
            \label{subfig:bouncingRBG}
        \end{subfigure}
        \begin{subfigure}[!tb]{0.49\linewidth}
            \centering
            \includegraphics[width=\linewidth]{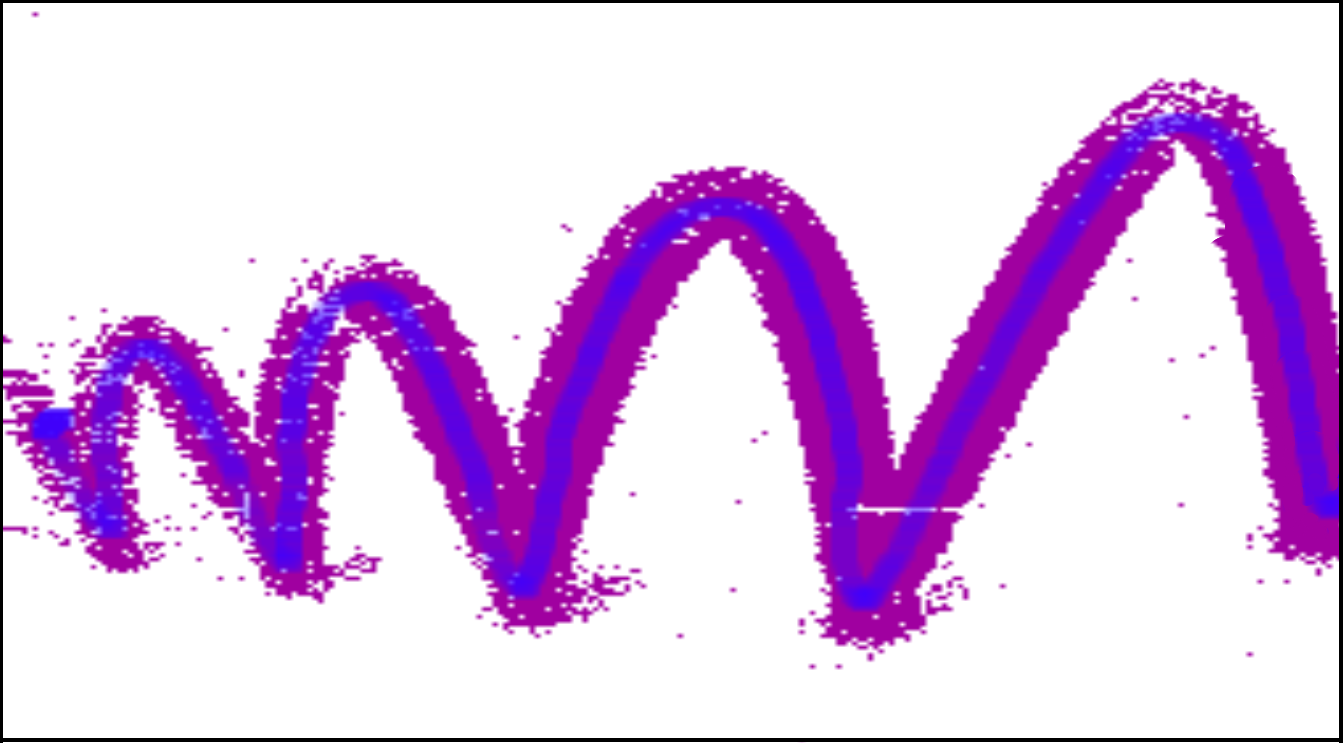}
            \caption{}
            \label{subfig:bouncingED}
        \end{subfigure}
  \caption{The trajectory of a bouncing ball comparing the sample points for (a) a traditional camera that samples at a fixed-rate in time, and (b) an event camera that samples with a fixed spatial sampling. The event camera output is shown with all the pixel-wise events (in purple) as well as the output of the trajectory tracker (in blue).}
  \label{fig:bouncingBall}
\end{figure}

To extend beyond what already exists in the field, we propose the use of \textit{event cameras}~\cite{ATIS}. Specifically, in this paper we analyse possible advantages in combining a classical deep learning strategy for sequence prediction with this type of sensors. Event cameras are a novel paradigm in machine vision that output a single \textit{event} for each pixel, but only when a change in the intensity of light falling on the pixel occurs. This happens when an object is moving in front of the camera, or at points of high contrast (e.g. edges of structure, edges of objects) when the camera itself moves. Such a technology acquires highly-compressed and low-latency (15 $\mu s$~\cite{ATIS}) visual trajectories of moving targets. The biggest conceptual difference when compared to a traditional camera is that an event camera samples a target trajectory based on its change in spatial value, i.e. the sampling of the vision signal is driven by the change in position. We therefore achieve a trajectory with samples at pixel, or sub-pixel, resolution, independently of the speed of the target. A comparison to a traditional camera, which samples at a fixed time period, is shown in Fig.~\ref{fig:bouncingBall}.

Event cameras are suited for dynamic tasks, such as camera pose estimation~\cite{6d0f_camera_relocalization,visual_inertial_odometry}, object tracking~\cite{arren_tracker}, and action recognition~\cite{ed_gesture_recog}. Visual learning has also been performed with event-data for classification~\cite{event-cnn1, event-cnn2}, ego-motion computation~\cite{event-driven_ego-motion}, depth estimation~\cite{event-driven_depth}, and vehicle steering prediction~\cite{event-driven_steering-angle1, ed_steering_prediction}. However, the majority of these works use convolutional neural networks (CNN) that do not inherently represent time, which is a very informative signal for event cameras. 


Recurrent Neural Networks (RNN) are designed to handle and learn patterns in sequential data. One of the most important RNN was introduced in 1990 by Elman~\cite{elman}, but, the work suffered from vanishing, exploding gradients~\cite{LSTM}. Long Short-Term Memory networks (LSTM)~\cite{LSTM} overcome this problem and are currently the state-of-the-art in the field of RNNs. 
The Encoder-Decoder - or Sequence-to-Sequence - model~\cite{seq2seq-text_transl} achieves state of the art results in several fields like Natural Language Processing~\cite{seq2seq-text_transl, seq2seq-speech_recog, seq2seq-image_capt} since it can map input sequences to output ones with different length. An Encoder-Decoder network has been combined also with a \textit{beam search} algorithm for predicting vehicle trajectories~\cite{seq2seq-vehicle_trajs} for self-driving cars. 

In this paper we investigate the integration of event-data with the LSTM architecture for learning to predict the trajectory of a bouncing ball focusing on the effect of different sampling strategies when combined with the learning architecture: \textit{fixed-rate} sampling, as performed in traditional cameras, compared with \textit{spatial} sampling, as is associated with event cameras. Experiments are performed using the neuromorphic iCub robot~\cite{Bartolozzi2011} equipped with the ATIS~\cite{ATIS} event camera. The ball bounces on a table in front of the robot and the robot, using the observed past ball positions, predicts the future trajectory for a given sequence. 

\section{Methodology}
\label{sec:methodology}

The system is comprised of the following components:
\begin{itemize}
    \item The ATIS~\cite{ATIS} event camera on the iCub robot;
    \item A target tracker that converts raw pixel events to the center of mass of the target along the trajectory;
    \item The trajectory sub-sampling strategy: either using a \textit{fixed-rate} sample or \textit{spatial} strategy; 
    \item An LSTM network, trained to predict the future trajectory of the target.
\end{itemize}

\paragraph{The ATIS camera}
Event cameras have a number of interesting properties that make them an intriguing option for pushing the visual systems of artificial agents.
Traditional cameras use a \textit{fixed-rate} sampling strategy in which all the pixels are queried for the light intensity at fixed intervals (e.g. 30 Hz). Therefore the trajectory of a target viewed with a traditional camera will correspond to different spatial positions at known time intervals, which leads to a trajectory such as in Fig.~\ref{subfig:bouncingRBG}.  For fast moving objects, fixed-rate sampling can lead to gaps in the trajectory as it is under-sampled, or motion blur.
The ATIS~\cite{ATIS} is a novel visual sensor with a resolution of $304\times240$ pixels. Each pixel produces an event when the amount of light changes beyond a threshold. Each pixel produces independent events with the following information: $< x_i,y_i,t_i,p_i >$, where $<x_i,y_i>$ is the pixel position on the sensor, $t_i$ is the time the event occurred with $\mu s$ resolution, and $p_i$ is the \textit{polarity} (whether an increase or decrease in light occurred). 
Therefore the readout is not bound by the time it takes to read the entire array. Instead pixels can emit events with a very short interval between them. The effect is that a moving target will trigger each and every pixel, consecutively, along its trajectory. The positions of target observation are triggered by a \textit{spatial} threshold (a change in pixel position), and the time at which a new observation occurs is \textit{asynchronous}.

\paragraph{Target tracking}
To simplify the input to the prediction system, the raw data from the ATIS is processed to extract the mean position of the target at any point in time. The target tracking is designed to reduced the effect of noise that could occur if the visual data is directly fed into the learning system, and focus on the effect of sampling strategies on the prediction. 
In this work, we implement a simple tracker with the simplifying assumption of a single moving target, as shown in Fig.~\ref{subfig:bouncingED}. The tracker accumulates events within a region-of-interest (ROI) of size $R$, and then updates the position of the ROI based on the average position of events within it. As the events move along the trajectory, so it does their mean position and the centre of the ROI. The tracker position is taken as the new centre of the ROI and is outputted along with the timestamp: $< x_{b_i},y_{b_i},t_{b_i}>$. Initialisation is performed by setting the ROI size $R$ equal to the entire image plane on the first update, and using the standard value thereafter.

\paragraph{Sub-sampling strategy}
The high temporal resolution of the event camera allows the trajectory to be sampled for every incoming event. The raw tracker data can therefore sample at over $1kHz$, but (as proposed in this paper) this raw signal can be sub-sampled to improve efficiency without compromising accuracy. Two sub-sampling strategies are proposed: in the \textit{fixed-rate} sampling we sample the signal at fixed time intervals, taking each sample after $F$ milliseconds; in the \textit{spatial} sampling we sample the signal when the spatial coordinates change by a fixed distance, taking each sample every $D$ pixels. The event camera enables very high sampling rates for both \textit{fixed-rate} and \textit{spatial} strategies. Note that the \textit{fixed-rate} strategy is more similar to a standard camera.

\begin{figure}
    \centering
    \centering
    \includegraphics[width=0.8\linewidth]{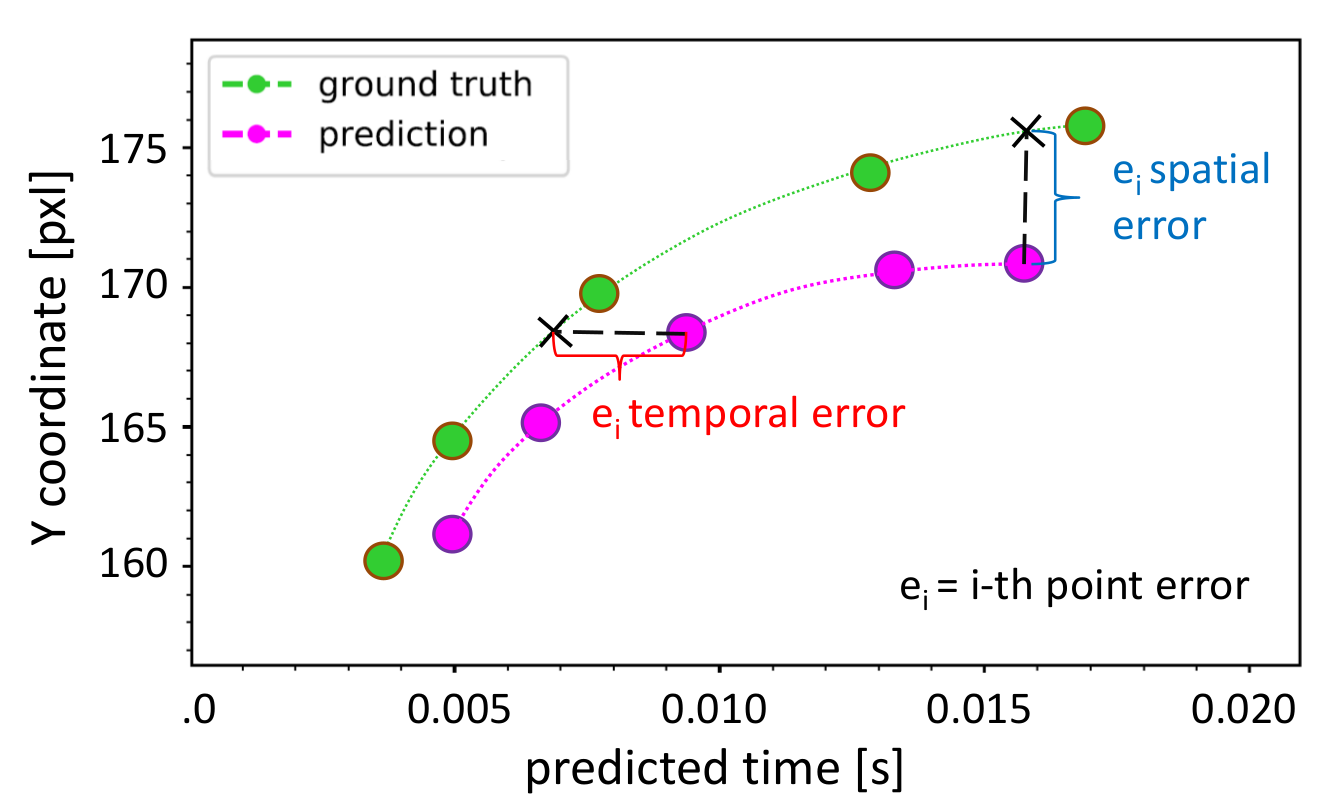}
    \caption{Spatial and temporal error component of the sequence-to-sequence error for a pair of prediction and ground truth.}
  \label{fig:error_definition}
\end{figure}

\paragraph{The LSTM network}

The model used in this work is a Sequence-to-Sequence LSTM architecture \cite{seq2seq-text_transl} consisting of two separate sub-networks. The Encoder takes the input data and encodes the information in a summary vector, represented by the state of the network after the input has been fed. This output becomes the initial state vector of the Decoder, that is queried to extract the information and produce the output sequence. The decoupling between the two stages allows to input sequences with a length $w_{in}$ different from the desired output sequence length $w_{out}$. This can be useful in tasks as machine translation or where the input and output physical domains differ, possibly requiring different representations. 
Our goal is to analyse how this architecture can be combined with event cameras for predicting the upcoming trajectory of a tracked object in an asynchronous fashion.
Commonly, RNNs are queried synchronously, limited by the working rate of devices like frame-based cameras. In our case, thanks to the event camera nature, the network can be queried asynchronously every time a new event is produced by the tracker. In such a way, not only it is exploited the information coming from previous queries, but also the temporal information contained within the events. To do so, the network is fed with a sequence comprising the spatial coordinates of the trajectory in the camera space and the time interval between the events in the sequence. 
The final outcome of the network is a prediction of the next $w_{out}$ points in space and time, which is continuously updated by querying the network for each new input. The prediction consists of future spatial coordinates and, for each point, its estimated time of arrival from the current instant. This spatio-temporal trajectory output could be helpful in robotics, allowing for accurate action planning, considering both spatial and temporal information.
The Encoder architecture considered has an input layer with 3 neurons, in order to accomodate the $(X, Y)$ spatial coordinates and the temporal interval $dT$ occurred from the previous input, and an output layer of 25 neurons. The Decoder architecture, on the contrary, consists of an input layer of 25 neurons to be initialised with the final Encoder state vector, and an output layer of 3 neurons outputting the future spatial position and its time of arrival.
The Sequence-to-Sequence model is first fed with the whole input sequence of $w_{in}$ points and, only then, the output sequence of $w_{out}$ point is generated at the same time.

\section{Experiments and Results}
\label{sec:experiments}

\begin{figure}
  \centering
        \includegraphics[width=\linewidth]{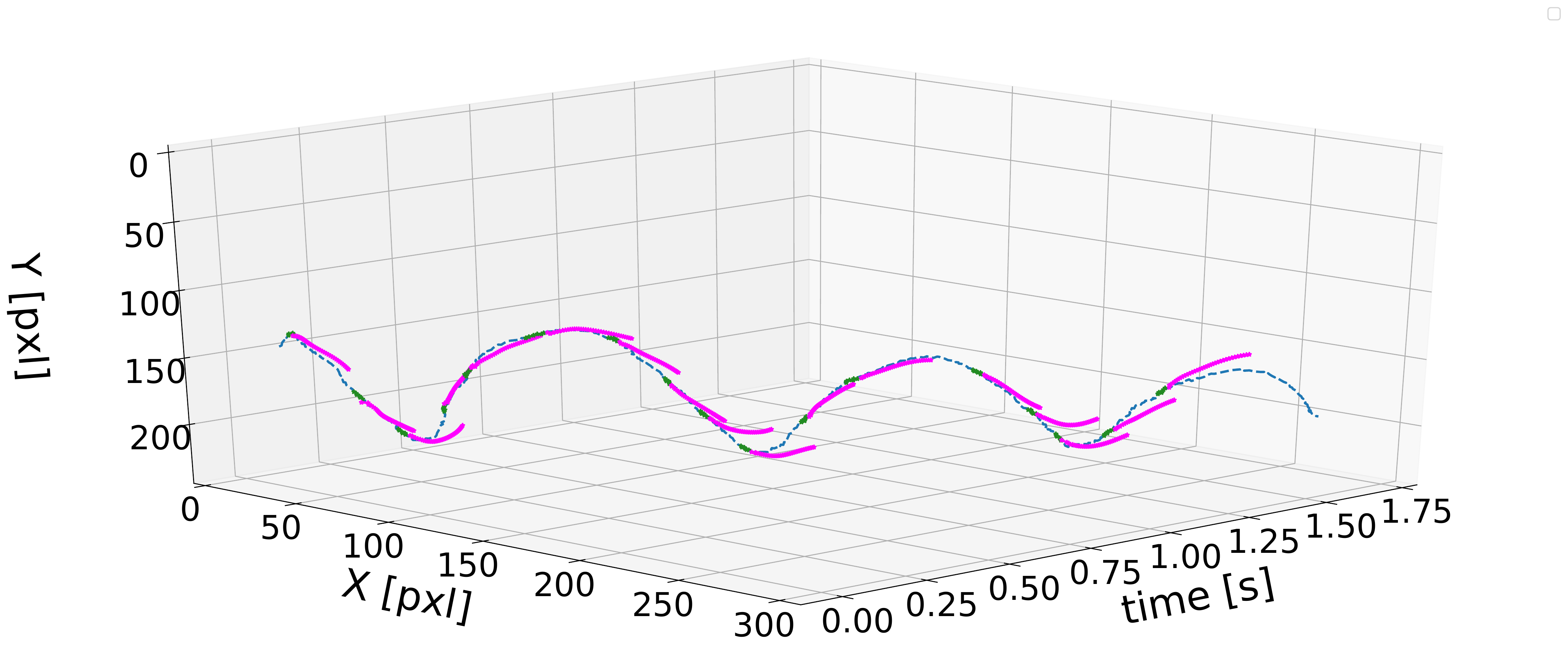}
  \caption{Complete example of a ball trajectory (in blue) with pairs of input (in green) and prediction (in purple) at different time instants.}
  \label{fig:whole_traj}
\end{figure}


A dataset was collected with 250 individual trajectories of the ball thrown in front of the camera and bouncing 2 or 3 times. The $<x_{b_i},y_{b_i},t_{b_i}>$ output of the tracker module was recorded. For independence to the direction of the trajectory, all datasets were flipped along the x-axis resulting in a total of 500 trajectories. The dataset was split into 470 training trajectories, 20 validation trajectories and 10 test trajectories. A trajectory example is shown in space-time in Fig.~\ref{fig:whole_traj}.

To choose the right Encoder-Decoder architecture, we started with a spatial sub-sampling of $D=2$ pixels and trained several models assuming different values for $w_{in}$ and $w_{out}$. Then, to compare the effect of the two sampling strategies, we chose suitable $F$ values for the \textit{fixed-rate} sampling and $D$ values for the \textit{spatial} sampling, to cover the same time intervals and looked at the test error. 



The Adam optimization algorithm~\cite{adam} was used with a learning rate of 0.01. Dealing with a regression problem, the Mean Squared Error (MSE) was adopted as loss function. No dropout was used and the networks were trained for 200 epochs on batches of size 128.

\begin{table}
\centering
\begin{tabular}{|| c| c ||}
\hline
Spatial delta D (pxl) & Mean+STD rate [Hz]\\ 
\hline\hline
-  & 357$\pm$88\\
2  & 182$\pm$44\\
4  & 92$\pm$22\\ 
6  & 63$\pm$14\\
8 & 48$\pm$11\\
10 & 39$\pm$9\\
12 & 33$\pm$7\\
\hline
\end{tabular}
\caption{Spatial delta sampling values with corresponding computational rate (mean and standard deviation).}
\label{table:samples}
\end{table}

 \subsection{Sequence-to-Sequence Characterisation}
 
To initially characterise some parameters of the LSTM, the prediction root-mean-square error (RMSE) error was calculated for prediction trajectories, according to trajectory spatial error and temporal error reported in to Fig.~\ref{fig:error_definition}. 
The RMSE is the mean error along the entire predicted length, and not only the final position.

\begin{figure}
    \centering
    \begin{subfigure}{\linewidth}
        \centering
        \includegraphics[width=.8\linewidth]{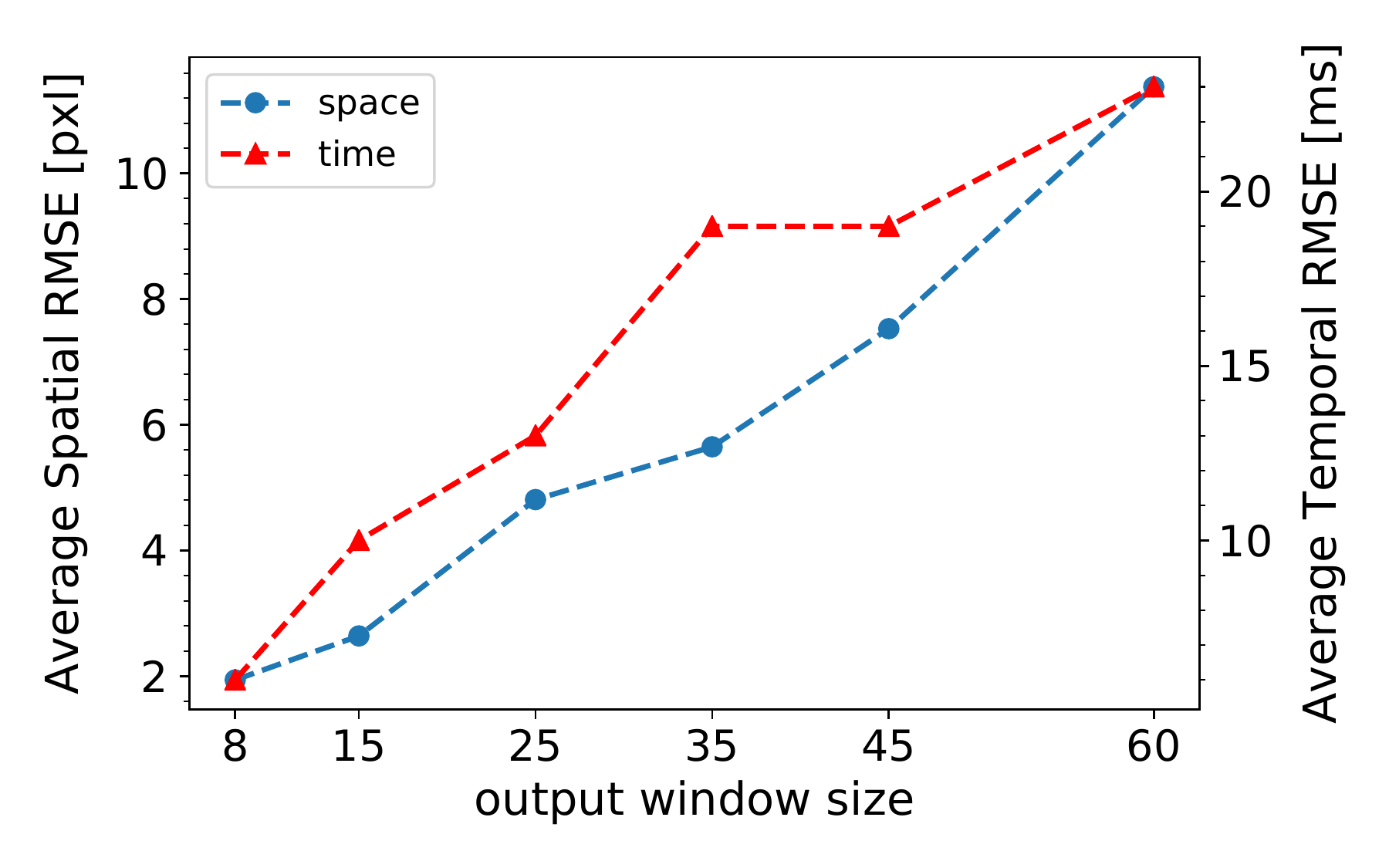}
        \caption{}
        \label{fig:RMSE_fixed_win}
    \end{subfigure}
    \begin{subfigure}{\linewidth}
        \centering
        \includegraphics[width=.8\linewidth]{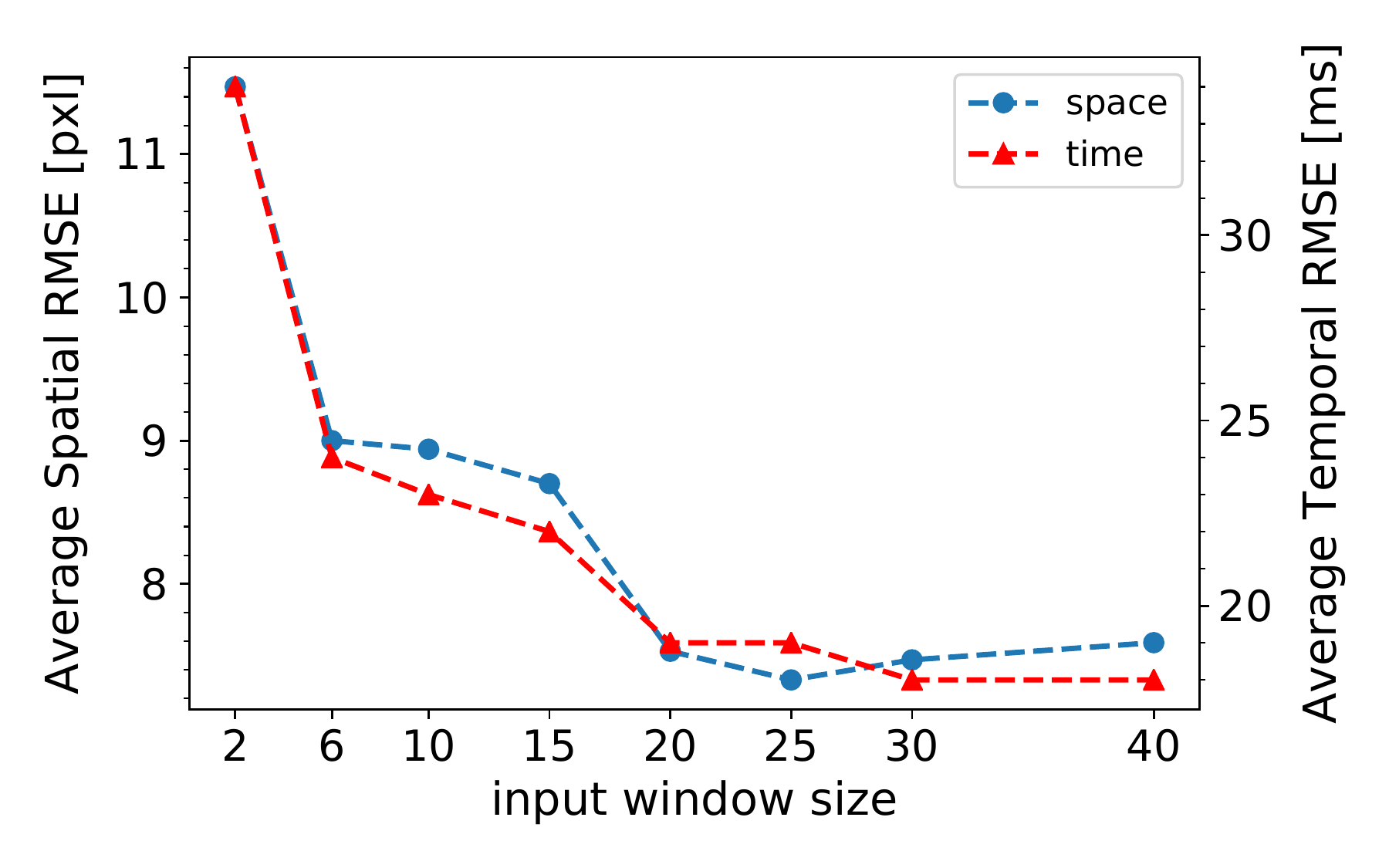}
        \caption{}
        \label{fig:RMSE_fixed_wout}
    \end{subfigure}
  \caption{Spatial and Temporal RMSE for different (a) $w_{out}$ values assuming $w_{in}=20$ and (b) $w_{in}$ values assuming $w_{out}=45$.}
  \label{fig:RMSE_winwout}
 \end{figure}

The error increases linearly with the output window length $w_{out}$, meant as number of points predicted in a single network interrogation, for both the spatial error and the temporal error, as shown in Fig.~\ref{fig:RMSE_fixed_win}. As the network tries to predict further ahead, the accuracy at which it can do so decreases. As the relationship is linear, there is no optimal point at which the network operates. The choice is therefore application specific, trading off the amount of time ahead that is needed, compared to the highest acceptable error.

The spatial error decreases the more data in the past is used, $w_{in}$. However, after a certain value, there is no real benefit in feeding more points, as shown in Fig.~\ref{fig:RMSE_fixed_wout}. In this case, a systematic operating point could be chosen to achieve the lowest error.

In the following, the output window length $w_{out}$ was set to be 45 and the input window length $w_{in}$ was set to 20.

\subsection{Sampling Strategy Comparison}

\begin{figure}[t]
  \centering
  \begin{subfigure}{\linewidth}
  \centering
  \includegraphics[width=0.9\linewidth]{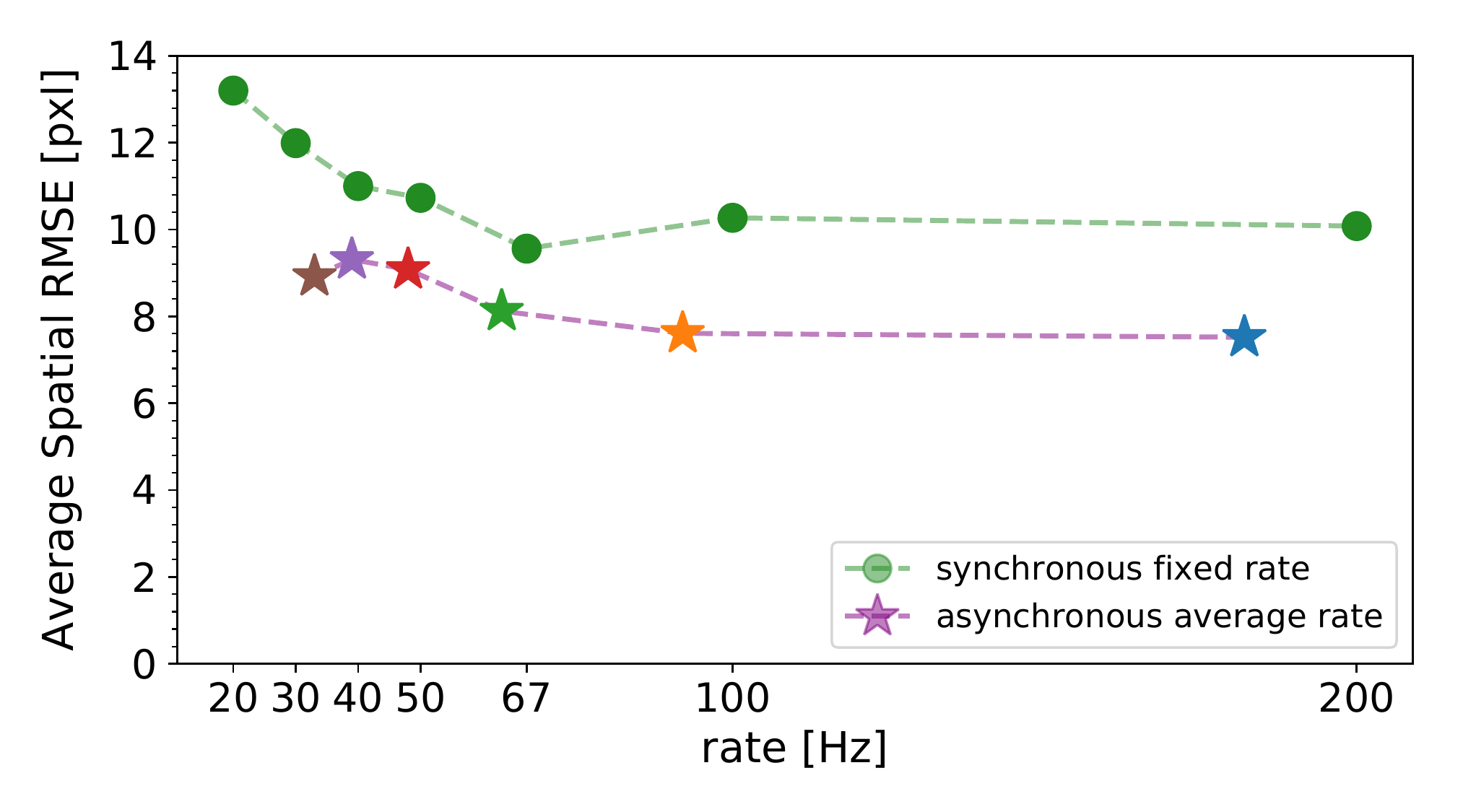}
  \caption{}
  \label{fig:srmse_different_rates}
  \end{subfigure}
  \begin{subfigure}{\linewidth}
  \centering
  \includegraphics[width=0.9\linewidth]{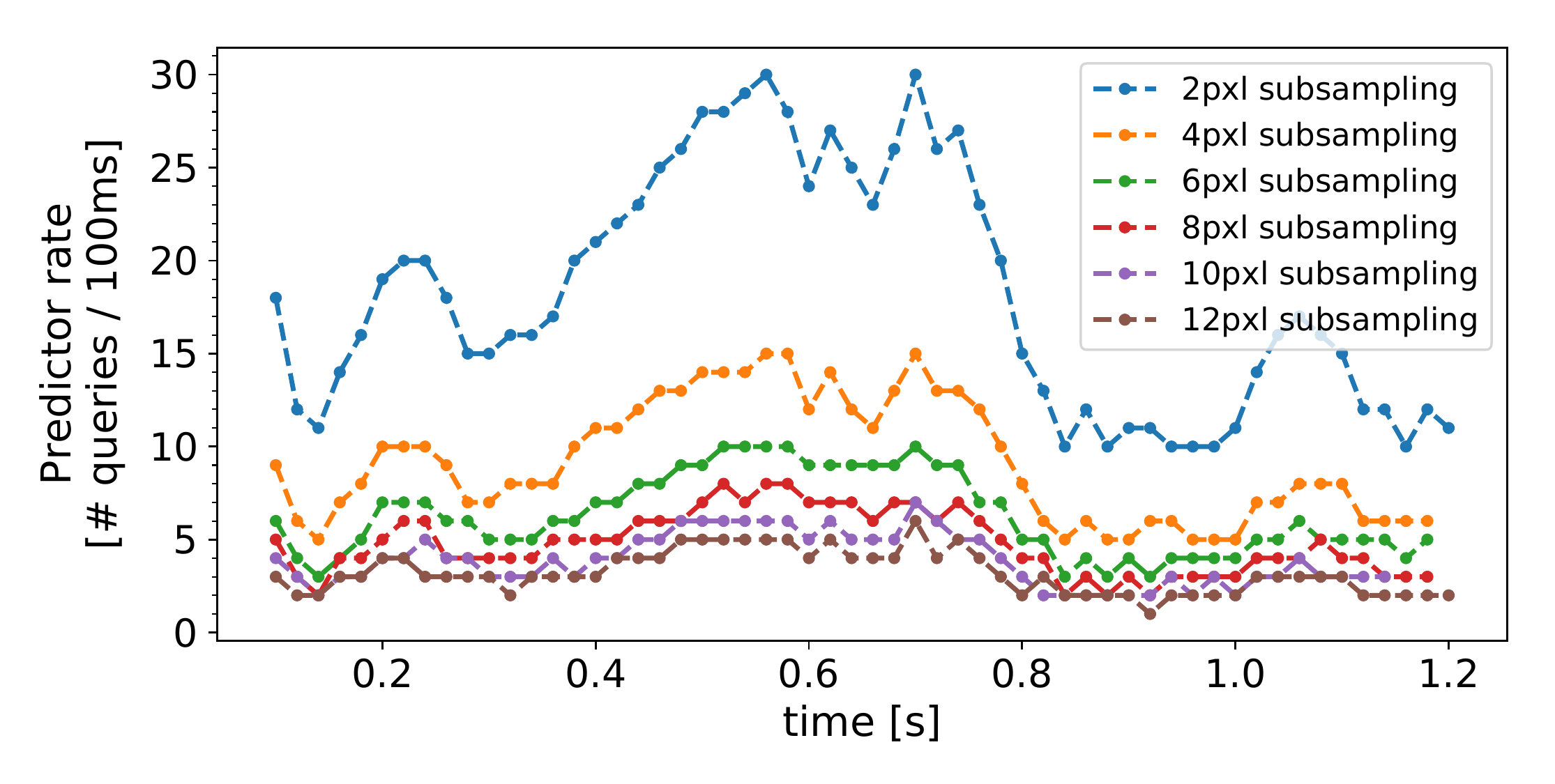}
  \caption{}
  \label{fig:predictor_rate}
  \end{subfigure}
  \caption{The (a) spatial RMSE for fixed-rate and spatial sampling, in which the mean sample rate of time is used to place the spatial sampling, and (b) the sample rate for spatial sampling over time for an example trajectory.}
\end{figure}

Secondly, a comparison between sampling strategies was made with respect to the prediction accuracy. Considering an input window of about $~90ms$ and an output window of about $~200ms$ defined by the previous windows length choice, we compare the performance of the two sub-sampling strategies on the same time intervals.

Fig.~\ref{fig:srmse_different_rates} shows the error for both sampling strategies. In order to perform a comparison, the mean sampling rate of the spatial was calculated (after sampling) and used as a comparison point to the fixed-rate sampling. Fixed-rate sampling showed an inverse relationship to error. A low rate results in a high error, and as the rate increases, the error decreases. However, the benefit of increased sample rate beyond 67 Hz is minimal. 

Spatial sampling instead shows a more linear decrease in error with respect to sampling rate. The error reaches a minimum value of 7.52 pixels for a spatial of 2 pixels, compared to a value of 10.08 pixels when using a fixed 200 Hz sampling rate. Importantly, the lower error is also achieved with less processing requirements as the average rate (and hence the total number of calculations) is lower for the 2 pixel spatial sampling.

The reason that the spatial sampling can give a lower error for an identical mean sampling rate is that it gives a varying sample rate in time, over the length of the dataset. In periods of fast and non-linear motion, where fine details are required to correctly trace the motion, the sample rate increases also enabling a more accurate prediction. On the other hand, when the target is moving slowly, the sample rate decreases and processing savings can be achieved. Fig.~\ref{fig:predictor_rate} shows an example sample rate (in time) for the spatial sampling, and it can be seen that for larger spatial deltas the sample rate becomes more constant across the dataset (a flatter profile). For sample rates without much variation, the error is much closer to that of fixed-rate sampling, as shown in Fig.~\ref{fig:srmse_different_rates}.

\section{Conclusions} 
\label{sec:conclusions}

We showed that LSTM networks could be adapted for event-based data, integrating the temporal component as an input to deal with asynchronous data. The network was characterised for a bouncing ball tracking task, and it was shown that only a small input window is required to achieve a low error. On the other hand, the output window can be extended for as long as necessary, with the trade-off that the error increases linearly. 

Asynchronous \textit{spatial} sampling outperforms \textit{fixed-rate} sampling and therefore we propose that such a strategy, combined with event cameras, is a promising approach for learning to predict trajectories for robotics tasks. For trajectories with variable speeds, more information can be captured by looking at the change in position with a high-resolution, rather than simply sampling at a high rate. Importantly \textit{spatial} sampling is made possible due to the event cameras, which allow asynchronous trajectory sampling directly triggered by change in the scene.



This study shows the advantage of spatial sampling in input data using event cameras and in the following visual processing, using asynchronous output events also in computational models such as the tracking, motivating a further quest in the use of event-driven perception coupled with LSTM architectures for fast vision tasks in artificial intelligence and robotics.
Depending on the exact constraints of the task, other solutions might be to have the network learn only the final position and time of a single trajectory, rather than the full path of the trajectory, allowing the robot to intercept the ball with the correct motion timing.
Future work will aim at improving the performance of the hereby presented architecture and evaluating other possible solutions, before moving to the robot control. 

\bibliographystyle{IEEEtran}
\bibliography{bibliography.bib}

\end{document}